\documentclass[letterpaper, 10 pt, conference]{ieeeconf}  %

\IEEEoverridecommandlockouts
\usepackage{times}
\usepackage{amsmath}
\usepackage{amsthm}
\usepackage{pifont}
\usepackage{amssymb}

\usepackage[squaren]{SIunits}
\usepackage{nicefrac}
\usepackage[utf8]{inputenc}
\usepackage{hyphenat}
\usepackage{xspace}
\usepackage{graphicx}
\usepackage{color}
\usepackage[ruled, linesnumbered, titlenotnumbered]{algorithm2e}
\usepackage{url}
\usepackage{footmisc}

\usepackage{dsfont}

\usepackage{mathtools}

\usepackage{glossaries}

\usepackage{relsize}
\usepackage{xcolor}
\usepackage{tikz}
\usetikzlibrary{decorations.pathreplacing, fit, calc, positioning, shapes,arrows,backgrounds}

\usepackage{booktabs}
\renewcommand{\arraystretch}{1.5}

\usepackage{setspace}

\theoremstyle{definition}
\newtheorem{definition}{Definition}

\definecolor{listinggray}{gray}{0.9}
\definecolor{lbcolor}{rgb}{0.9,0.9,0.9}
\definecolor{darkgreen}{rgb}{0.0, 0.2, 0.13}

\usepackage{todonotes}

\usepackage{adjustbox}
\usepackage{array}

\newcolumntype{R}[2]{%
    >{\adjustbox{angle=#1,lap=\width-(#2)}\bgroup}%
    l%
    <{\egroup}%
}
\newcommand*\rot{\multicolumn{1}{R{45}{1em}}}%

\usepackage{textcomp}
\usepackage{listings}
\makeatletter%
\let\l@ngrel@x\protected\provide@command{\nbhyp}{%
  \nobreak\mbox{-}\nobreak\hskip\z@skip}
\makeatother%

\newcommand{\removelatexerror}{\let\@latex@error\@gobble}

\newcommand{\rospkg}[1]{\texttt{#1}}

\newcommand{\pythonclass}[1]{\lstinline[language=Python]{#1}}

\newcommand{\pythonattr}[1]{\lstinline[language=Python]{#1}}

\newcommand{\rosmsg}[2][]{%
  \ifthenelse{ \equal {#1} {}}
  {}
  {\textcolor{black!75}{\rospkg{#1/}}\hspace{0pt}}%
  \texttt{#2}%
}

\newcommand*{\varname}[1]{\mathord{\mathit{#1}}}
\newcommand{\btstates}{\mathcal{S}}
\newcommand{\btactions}{\mathcal{A}}
\newcommand{\btgraph}{\mathcal{G}_T}
\newcommand{\btworld}{\mathcal{W}}

\newcommand{\btdataalph}{\Sigma}

\newcommand{\btparams}{\mathcal{P}}
\newcommand{\btparamtypes}{\mathbf{T}}
\newcommand{\btnodes}{\mathcal{N}_T}
\newcommand{\btedges}{\mathcal{E}_T}
\newcommand{\btenv}{\mathit{Env}_T}
\newcommand{\btorderfunc}{\mathcal{o}}
\newcommand{\btdef}{$ \btgraph{}(\btnodes{}, \btedges{}, \btorderfunc{}) $}
\newcommand{\btenvdef}{$\btenv{} = (\btgraph{}, \datagraph{}, \btdataalph{}, \btworld{})$}

\newcommand{\paramkind}{k}
\newcommand{\paramtype}{t}

\newcommand{\wiringsource}{p_s}
\newcommand{\wiringtarget}{p_t}

\newcommand{\subtreegraph}{\mathcal{G}_S}
\newcommand{\subtreenodes}{\mathcal{N}_S}
\newcommand{\subtreeedges}{\mathcal{E}_S}
\newcommand{\subtreeorderfunc}{\mathcal{o}_S}
\newcommand{\subtreedef}{$ \subtreegraph{} = (\subtreenodes{}, \subtreeedges{}, \subtreeorderfunc{}) $}
\newcommand{\subtreeenv}{\mathit{Env}_S}

\newcommand{\btutil}{\mathcal{U}}
\newcommand{\btutildef}{$\btutil{} = \mathbb{R} \cup \left\{ \text{\notavailablemark{}}, ? \right\}$}

\newcommand{\datagraph}{\mathcal{G}_D}
\newcommand{\datanodes}{\mathcal{N}_D}
\newcommand{\dataedges}{\mathcal{E}_D}
\newcommand{\datadef}{$ \datagraph{}(\datanodes{}, \dataedges{}) $}

\newcommand{\minfailcost}{c^{f\protect\vphantom{f}}_{\protect\vphantom{i}\text{min}}}
\newcommand{\maxfailcost}{c^{f\protect\vphantom{f}}_{\protect\vphantom{i}\text{max}}}
\newcommand{\minsucccost}{c^{s\protect\vphantom{f}}_{\protect\vphantom{i}\text{min}}}
\newcommand{\maxsucccost}{c^{s\protect\vphantom{f}}_{\protect\vphantom{i}\text{max}}}

\newcommand{\crossmark}{\ding{55}}
\newcommand{\notavailablemark}{\ding{53}}

\definecolor{fzi-green}{RGB}{13,114,73}

\definecolor{succeeded}{HTML}{1b5e20}
\definecolor{failed}{HTML}{b71c1c}
\definecolor{running}{HTML}{6A1B9A}%
\definecolor{idle}{RGB}{0,0,0}
\definecolor{uninitialized}{HTML}{455a64}
\definecolor{error}{HTML}{CC4700}
\definecolor{shutdown}{HTML}{1a237e}

\tikzset{
  treenode/.style = {draw=idle, align=center, inner sep=0pt, thick, text centered,
    font=\sffamily, text=black},
  btcondition/.style = {treenode, ellipse, inner sep=4pt, draw=black},
  btaction/.style = {treenode, rectangle, inner sep=4pt, draw=black},
  btdecorator/.style = {treenode, chamfered rectangle, inner sep=4pt, draw=black},
  btnode/.style = {treenode, rectangle, inner sep=4pt, black, draw=black, text height=0.8em, text width=1.2em},
  round/.style={rounded corners=1.5mm,minimum width=1cm,inner sep=2mm,above right,draw},
  bt-idle/.style={draw=idle, thick},
  bt-uninitialized/.style={draw=uninitialized, thick},
  bt-succeeded/.style={draw=succeeded, thick},
  bt-failed/.style={draw=failed, thick},
  bt-running/.style={draw=running, thick},
  bt-error/.style={draw=error, thick},
  bt-shutdown/.style={draw=shutdown, thick}
}

\tikzset{fontscale/.style = {font=\relsize{#1}}}

\newcommand{\sequence}[1]{node [btnode] (#1) {$\rightarrow$}}
\newcommand{\fallback}[1]{node [btnode] (#1) {$?$}}

\newacronym{AI}{AI}{Artificial Intelligence}
\newacronym{BT}{BT}{Behavior Tree}
\newacronym{GUI}{GUI}{Graphical User Interface}
\newacronym{FSM}{FSM}{Finite State Machine}
\newacronym{HSFM}{HFSM}{Hierarchical Finite State Machine}
\newacronym{HTN}{HTN}{Hierarchical Task Network}
\newacronym{ROS}{ROS}{Robot Operating System}
\newacronym{FZI}{FZI}{FZI Research Center for Information Technology}
\newacronym{STRIPS}{STRIPS}{Stanford Research Institute Problem Solver}
\newacronym{UAV}{UAV}{Unmanned Aerial Vehicle}
\newacronym{MCA2}{MCA2}{Modular Controller Architecture 2}
\newacronym{JSON}{JSON}{JavaScript Object Notation}
\newacronym{DRC}{DRC}{DARPA Robotics Challenge}
\newacronym{RPC}{RPC}{Remote Procedure Call}
\newacronym{RAFCON}{RAFCON}{RMC advanced flow control}
\newacronym{RMC}{RMC}{Robotics and Mechatronics Center}
\newacronym{DLR}{DLR}{German Aerospace Center}
\newacronym{HPFD}{HPFD}{Hierarchical, Parallel Finite state machine with Data flows}
\newacronym{BLE}{BLE}{Broadcast of Local Eligibility}
\newacronym{MVC}{MVC}{Model-View-Controller}
\newacronym{YARP}{YARP}{Yet Another Robot Platform}
\newacronym{LTL}{LTL}{Linear Temporal Logic}
\newacronym{URI}{URI}{Uniform Resource Identifier}
\newacronym{URDF}{URDF}{Unified Robot Description Format}
\newacronym{STDR}{STDR}{Simple Two-Dimensional Robot}
\newacronym{PLEXNAV}{PLEXNAV}{PLanetary EXploration and NAVigation}
\newacronym{hollie}{HoLLiE}{the House of Living Labs intelligent Escort}

\title{\LARGE \bf
Distributed Behavior Trees for Heterogeneous Robot Teams
}

\author{%
  Georg Heppner$^{1}$ \and Nils Berg$^{1,2}$ \and David Oberacker$^{1}$ \and Niklas Spielbauer$^{1}$\and A. Roennau$^{1}$ \and R. Dillmann$^{1}$%
  \thanks{$^{1}$\ Department of Interactive Diagnosis and Service
    Systems (IDS), FZI Research Center for Information Technology,
    Haid-und-Neu-Straße 10--14, 76131~Karlsruhe, Germany.
    }%
      \thanks{$^{2}$\ Now at: Intrinsic Innovation GmbH,
    ABC-Straße 19, 20354~Hamburg, Germany.}%
}

\newcommand\copyrighttext{%
    \footnotesize \textcopyright 2023 IEEE. Personal use of this material is permitted.
    Permission from IEEE must be obtained for all other uses, in any current or future
    media, including reprinting/republishing this material for advertising or promotional
    purposes, creating new collective works, for resale or redistribution to servers or
    lists, or reuse of any copyrighted component of this work in other works.}
\newcommand\copyrightnotice{%
    \begin{tikzpicture}[remember picture,overlay]
        \node[anchor=south,yshift=10pt] at (current page.south) {\fbox{\parbox{\dimexpr\textwidth-\fboxsep-\fboxrule\relax}{\copyrighttext}}};
    \end{tikzpicture}%
}

\begin{document}

\maketitle

\thispagestyle{empty}
\pagestyle{empty}

\copyrightnotice

\begin{abstract}

Heterogeneous Robot Teams can provide a wide range of capabilities and therefore significant benefits when handling a mission.
However, they also require new approaches to capability and mission definition that are not only suitable to handle heterogeneous capabilities but furthermore allow a combination or distribution of them with a coherent representation that is not limiting the individual robot.
\glspl{BT} offer many of the required properties, are growing in popularity for robot control and have been proposed for multirobot coordination, but always as separate \glspl{BT}, defined in advance and without consideration for a changing team.
In this paper, we propose a new \acrlong{BT} approach that is capable to handle complex real world robotic missions and is geared towards a distributed execution by providing built in functionalities for cost calculation, subtree distribution and data wiring.
We present a formal definition, its open source implementation as \emph{\textbf{ros\_bt\_py}} library and experimental verification of its capabilities.
\end{abstract}

\section{INTRODUCTION}
\glsresetall
As robots and the tasks they are expected to perform are becoming more complex, the need for flexible and extendable mission control systems and definitions has steadily increased.
\glspl{HSFM} or \glspl{HTN} are often used to define the missions as a hierarchical combination of robot capabilities but in recent years \glspl{BT} have become a popular alternative.
\glspl{BT} offer a compact, yet intuitive, representation for missions as well as individual capabilities, that combines behavior and control flow definitions and is particularly well suited for modular reuse and hierarchical decomposition while providing a high reactivity.

With an increasing availability of heterogeneous robots and their use in a shared environment there is a growing need for a capability and mission definition that is not only suitable to handle the heterogeneous capabilities of the various systems but furthermore allows a combination or  distribution of them with a coherent representation.
We believe that Behavior Trees are the perfect approach to fill this need and propose an extension to previous definitions that can leverage them for distributed task execution.

\section{STATE-OF-THE-ART}
\begin{table}[b]
    \caption{Comparison of different Behavior Tree implementations for robotics.}
    \label{tab:implementations}
    \resizebox{0.8\textwidth}{!}{\begin{minipage}{\textwidth}
            \begin{tabular}{l | c c c c c c c c}
                \rot{\textbf{Name}} & \rot{\textbf{Reactivity}} & \rot{\textbf{Arguments}} & \rot{\textbf{Black Board}} & \rot{\textbf{Data Wirings}} & \rot{\textbf{Async}} & \rot{\textbf{GUI}} & \rot{\textbf{ROS}} & \rot{\textbf{Distributed Exec.}}\\
                \hline
                BehaviorTree.cpp\footnote{\url{https://github.com/BehaviorTree/BehaviorTree.CPP}} & \checkmark & \checkmark & \checkmark & \crossmark & \checkmark & \checkmark & \checkmark & \crossmark \\
                Py\_trees\footnote{\url{https://github.com/splintered-reality/py_trees}} & \crossmark & \crossmark & \checkmark & \crossmark & \crossmark & \checkmark & \checkmark & \crossmark \\
                CoSTAR\footnote{\cite{Guerin15}} & \crossmark & \checkmark & \checkmark & \crossmark & \crossmark & \crossmark & \crossmark & \crossmark \\
                BrainTree\footnote{\url{https://github.com/arvidsson/BrainTree}} & \checkmark & \checkmark & \checkmark & \crossmark & \crossmark & \crossmark & \crossmark & \crossmark \\
                go-behave\footnote{\url{https://github.com/askft/go-behave}} & \crossmark & \checkmark & \checkmark & \crossmark & \crossmark & \crossmark & \crossmark & \crossmark \\
                ActionFW\footnote{\url{https://bitbucket.org/brainific/action-fw}} & \checkmark & \crossmark & \crossmark & \crossmark & \crossmark & \crossmark & \crossmark & \crossmark \\
                \hline
                \textbf{ros\_bt\_py}\footnote{\url{https://github.com/fzi-forschungszentrum-informatik/ros_bt_py}\label{fn:ros_bt_py}} & \checkmark & \checkmark & \crossmark & \checkmark & \checkmark & \checkmark & \checkmark & \checkmark \\
            \end{tabular}
    \end{minipage}}
\end{table}
Behavior Trees stem from the AI development in video games \cite{Isla05, Champandard2007} and have since matured to a standard in game engines.
\"Ogren proposed their use as a Hybrid Dynamical System (HDS) to switch controllers of UAVs \cite{Ogren2012}.
Marzinotto and Colledanchise continue this work in \cite{ICRA14Marzinotto} with a comparative overview of much of the previous works on \glspl{BT} and a formal model to use \glspl{BT} for robotics which is widely cited as inspiration for further works.
In their work, the authors propose \glspl{BT} as a mediating layer between low-level controllers interacting with the robots hardware and a high level planning layer using \gls{LTL}.
Later work bridges the gap between planning and execution by showing how to synthesize \glspl{BT} from \gls{LTL} \cite{Colledanchise17}, directly using \glspl{BT} as part of the temporal logic \cite{Martens2018} and back chaining \cite{Colledanchise19}.
Colledanchise and \"Ogren also provide a \acrfull{BT} implementation \cite{TRO17Colledanchise} and a detailed overview of \glspl{BT} for robotics \cite{BTBook} as well as a more recent implementation overview \cite{9448466}.
Iovino et al. \cite{IOVINO2022104096}, and Ghzouli et al. \cite{10.1145/3426425.3426942} provide in-depth surveys on the further recent usage of \glspl{BT} in robotics and engineering.
Table~\ref{tab:implementations} gives an overview of available open-source implementations of \glspl{BT} for robotics.
Other fields where \glspl{BT} have been successfully applied include medical operations \cite{Hu15}, swarm robotics \cite{9636460, neupane_learning_2019} and manufacturing \cite{Guerin15, 7989070, sidorenko_using_2022}.

Some interesting additions to \gls{BT} definitions which this work builds upon were made by Kl\"ockner \cite{klockner2013behavior} who introduced new states to \gls{BT} nodes and deliberate entry and exit hooks, enabling long running tasks and Shoulson et al. \cite{parameterizing_BTs} who introduced parameters for \glspl{BT} to increase their versatility in reuse.

\begin{figure}[htb]
  \centering
  \includegraphics[width=\columnwidth]{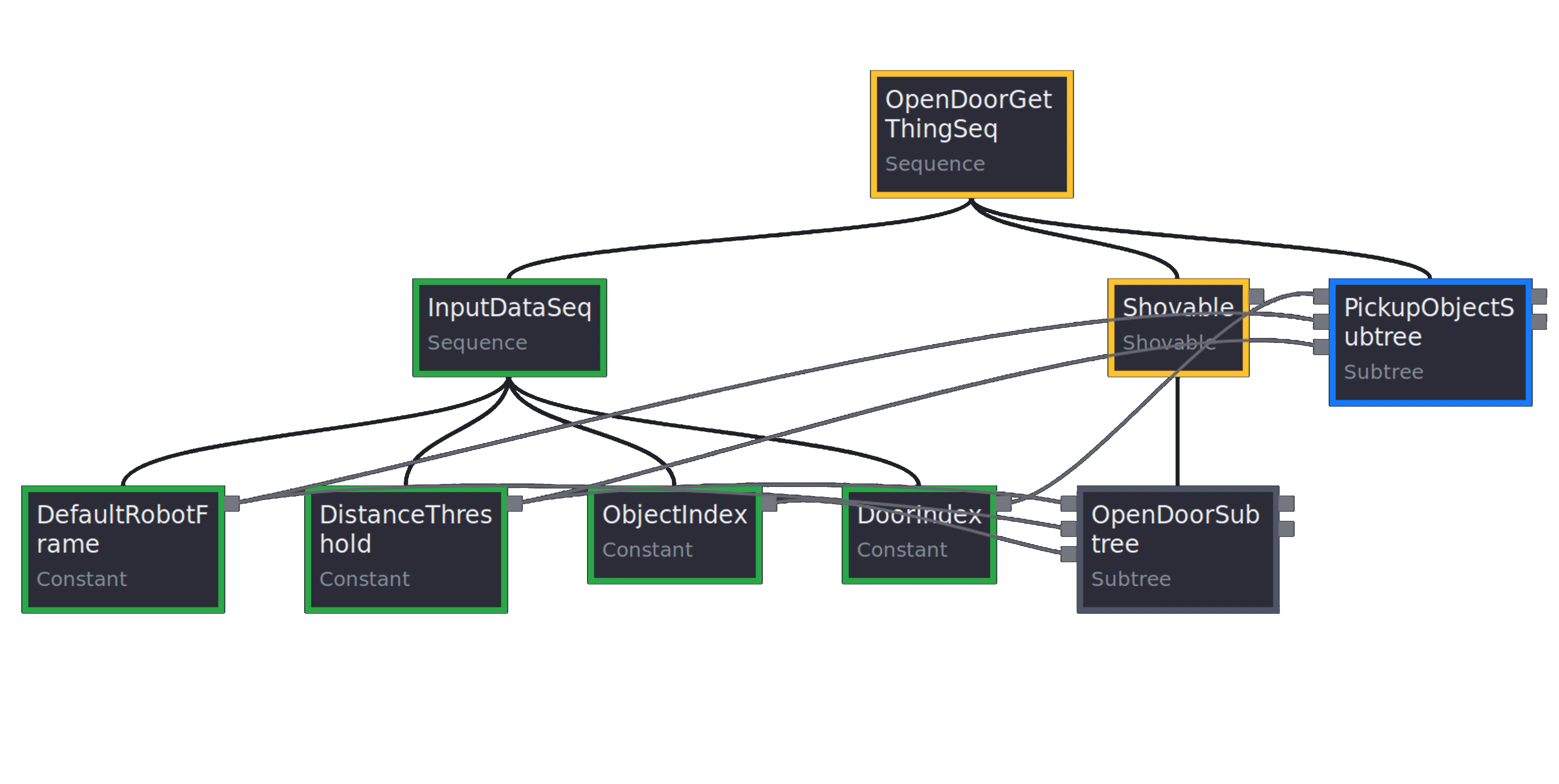}
  \caption{\gls{BT} used for the multirobot experiment using the proposed architecture. The open door subtree is pushed behind a shovable decorator, marking it as a part for possible remote execution. The data wirings are transparently forwarded.}
    \label{fig:multirobot_experiment_tree}
\end{figure}

\glspl{BT} have also been shown to be used for Multi Robot Systems (MRS).
Colledanchise et al. \cite{ISR16Colledanchise} present a methodology to transform a local \gls{BT} into a distributed one by constructing a global, a local and an assignment tree that are running in parallel.
The global tree contains the original mission but replaces sequences with parallel execution whenever possible, the local tree checks which task is assigned and executes it while the task assignment tree solves an optimization problem to assign tasks to agents.
Jeong et al. \cite{https://doi.org/10.48550/arxiv.2201.10918} present an approach for navigating a team of robots using ROS 2 (Robot Operating System) for communication and \glspl{BT} for mission control.
They provide \gls{BT} nodes for sending and receiving data via ROS topics, allowing for the coordination between robots.
A central task scheduler \gls{BT} makes use of these topic nodes to dispatch navigation goals to individual robots, with each running their individual \gls{BT} responsible for navigation and possible recovery behaviors.
Both approaches use the inherent flexibility of \glspl{BT} to separate the overall mission into modular parts that are executed individually, but do so in a predefined way that limits the approaches adaptability and flexibility.
The team size and participants need to be known in advance and only pre-defined mission information or synchronization information is exchanged during runtime while the \glspl{BT} remain static.
Venkata et al.\cite{venkata_kt-bt_2022} propose Knowledge-Transfer Behavior Trees (KT-BTs), which allow robots to teach new skills to teammates.
Thus, individual team member become more capable as new skills are presented to the team, increasing redundancy and improving the overall teams capabilities.

This work proposes a unified Behavior Tree approach for heterogeneous robot teams that can adaptively distribute parts of the \gls{BT} among the team to facilitate their local capabilities, while retaining an overall mission definition and collecting relevant data and execution results in a transparent way.
The work extends the previous works on \glspl{BT} for robotics to create a system that is geared towards real world usage by introducing concepts such as ticking, unticking, parameters and a data graph while introducing the ability to executing parts of a tree on a remote entity for easy task distribution and coordination of heterogeneous robot teams.

The paper is structured as follows: After this introduction with related work, section \ref{sec:Behavior_Tree_Definition} introduces our general Behavior Tree definition which is then subsequently extended with long running asynchronous tasks (\ref{sec:long_running_tasks}), a data graph (\ref{sec:data_graph}), utility caluclations (\ref{sec:utility_functions}) and shovables (\ref{sec:shoving}).
Section \ref{sec:Implementation} then illustrates and compares the implementation and shows its practical realization with an example before section \ref{sec:conclusion} summarizes the contribution and the next research steps.
\begin{figure}[hb]
  \centering
  \begin{tikzpicture}[->,>=stealth',level/.style={sibling distance = 5cm/#1, level distance = 1.5cm}]
    \node [btnode] {$?$}
      child { node [btcondition] {Have Ball?}}
      child { \sequence{sel}
        child { node [btaction] {Detect Ball}}
        child { node [btaction] {Pick Up Ball}}
      }
      ;
  \end{tikzpicture}
  \caption[A simple \gls{BT} to detect and pick up a ball]{Example of a simple \gls{BT} -- the top-most node is the root. The tree first checks whether \emph{Have Ball?} is true. If so, the \emph{Fallback} succeeds and the rest of the tree is not executed. If not, the \emph{Sequence} executes its children \emph{Detect Ball} and \emph{Pick Up Ball}, succeeding if both succeed, and failing otherwise.}

  \label{fig:concept-bt-example-1}
\end{figure}
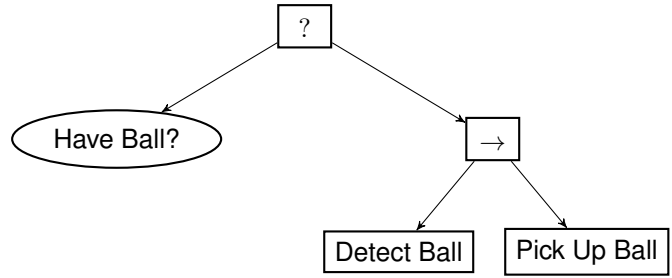

\section{Behavior Tree Definition}
\label{sec:Behavior_Tree_Definition}
We propose a complete Behavior Tree approach that extends the designs previously presented for robotic use.
The basic definition closely follows approaches such as by Marzinotto \cite{ICRA14Marzinotto} and Colledanchise \cite{BTBook}.
This general definition is then extended to support the following key aspects:
\begin{itemize}
\item{\textbf{Long running asynchronous tasks}}\\
This is a key requirement to enable long running action for robotics such as movements of a mobile base
\item{\textbf{Modular, parameterizable Nodes and data wiring}}\\
This is required to use \glspl{BT} to define complex robotic missions and avoid redundant nodes or trees.
\item{\textbf{Utility calculation and aggregation}}\\
When distributing Behavior Trees to multiple robots, a metric is required to calculate on the execution costs.
\item{\textbf{Shoving and Slots for distributed execution}}\\
Encapsulating behavior into subtrees is the core principle for efficient reuse.
Shovables extend this principle by executing a tree across multiple systems to enable task distribution for a robot team.
\item{\textbf{Extensible architecture geared towards ROS}}\\
The approach needs to be extensible to adapt the ever changing requirements of robotics.
In particular, ROS as de-factor standard for many robots, should be supported.
\end{itemize}

\begin{definition}[Node]
  \label{def:node}
  A \emph{Node} $n$ is an entity having a State
  \[
    \begin{split}
  \varname{state}(n) \in \btstates{} = \{
   & \text{uninitialized}, \text{error}, \text{idle}, \text{succeeded},\\
   & \text{failed}, \text{running},\text{shutdown} \}
    \end{split}
  \]
  and a maximum number of children $\varname{maxChildren}(n) \in \mathopen[0, \infty \mathclose)$.
  A Node's State can be changed by an $\varname{update}$ function (see definition \ref{def:btenv}), depending on the \emph{Action}
    \[
    a \in \btactions{} = \{
    \text{setup},
    \text{tick},
    \text{untick},
    \text{reset},
    \text{shutdown}
    \}
    \]
\end{definition}
Figure \ref{fig:node-state-transitions} shows the possible transitions for each action.
Which transition is taken upon receiving a given Action (if there are multiple possibilities) depends on the Node.
\begin{definition}[Node Classes]
  Nodes can be categorized into three basic \emph{Node Classes}:
  \emph{Flow Control}, \emph{Decorators} and \emph{Leaves}.
  For any Node $n$:
  \[
  \varname{nodeClass}(n) =
  \begin{cases}
    \text{Leaf} & \text{if} \  \varname{maxChildren}(n) = 0,\\
    \text{Decorator} & \text{if} \  \varname{maxChildren}(n) = 1,\\
    \text{Flow Control} & \text{if} \  \varname{maxChildren}(n) > 1
  \end{cases}
  \]
\end{definition}
Flow Control nodes change the way the \emph{tick} is propagated in the tree.
Commonly used ones are the \emph{sequence, fallback} and \emph{parallel} nodes (see \cite{ICRA14Marzinotto} for an algorithmic definition).
\begin{figure}[tb]
  \centering
  \scalebox{0.55}{
      \begin{tikzpicture}[->,>=stealth',shorten >=1pt,auto,node distance=4cm, nodelabel/.style={font=\relsize{+1}}]
      \tikzstyle{state}=[rectangle,draw]
        \node[state, double] (A)                    {uninitialized};
        \node[state]         (B) [right of=A] {idle};
        \node[state, align=center]         (D) [right of=B] {running,\\succeeded,\\failed};
        \node[state]         (E) [below right of=D] {shutdown};
        \node[state]         (C) [above right of=D] {paused};

        \begin{scope}[on background layer]
         \path
          (A) edge              node [nodelabel] {setup} (B)
          (A) edge [bend right] node [nodelabel, swap] {shutdown} (E)
          (B) edge [bend right] node [nodelabel, swap] {shutdown} (E)
          (B) edge              node [nodelabel] {tick} (D)
          (B) edge [loop above] node [nodelabel, align=center] {untick,\\reset} (B)
          (D) edge [loop right] node [nodelabel] {tick} (D)
          (D) edge              node [nodelabel, swap] {untick} (C)
          (D) edge [bend right] node [nodelabel, swap, align=center, above right] {untick,\\reset} (B)
          (D) edge              node [nodelabel] {shutdown} (E)
          (C) edge [bend right] node [nodelabel, swap] {tick} (D)
          (C) edge [loop right] node [nodelabel] {untick} (C)
          (C) edge [bend right] node [nodelabel, swap] {reset} (B)
          (C) edge [bend left]  node [nodelabel] {shutdown} (E)
          (E) edge              node [nodelabel] {setup} (B)
          (E) edge [loop right] node [nodelabel] {shutdown} (E);
        \end{scope}
      \end{tikzpicture}
  }
  \caption[The possible State transitions when applying an Action to a Node.]{The possible State transitions when applying an Action to a Node. For added clarity, the States \emph{succeeded}, \emph{failed} and \emph{running} have been combined.}
  \label{fig:node-state-transitions}
\end{figure}
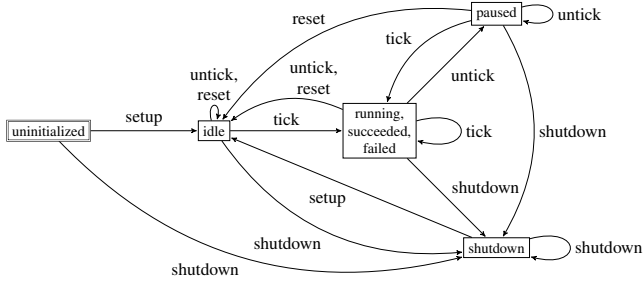
\begin{definition}[\glsentrylong{BT}]
  \label{def:bt}
  A \glsentryfull{BT} is an \emph{ordered tree}~\cite[p.~573]{stanley11}, described here as a graph $ \btgraph{}(\btnodes{}, \btedges{}, \btorderfunc{}) $, where $ \btnodes{} $ is the set of Nodes in the \gls{BT}, $ \btedges{} $ the set of directed edges $ (p, c)$ with $p,c \in \btnodes{}$ and $p \neq c $, and $ \btorderfunc{} \colon \btedges{} \to \mathbb{N} $ a function mapping each \emph{Edge} $(p, c)$ to its position among the children of $p$.

  To do this, $\btorderfunc{}$ needs to fulfill the following condition:
  \[
    \btorderfunc{}\left((p, c)\right) \neq \btorderfunc{}\left((p, x)\right) \quad %
    , \forall x \in \btnodes{} \setminus \{p, c\}
  \]

That is, no two children of the same parent may have the same position.
\end{definition}
\begin{definition}[Root of a \gls{BT}]
  Since a \gls{BT} is an \emph{ordered tree}, which is a kind of \emph{rooted tree}, there is guaranteed to be a single \emph{Root}.
  That is, in every \gls{BT} \btdef{}, there exists a \emph{Root} $r \in \btnodes{}$ so that $(n, r) \notin \btedges{}, \forall n \in \btnodes{} \setminus \{ r \} $.

  The State of the \emph{Root} $\varname{state}(r)$ is also called the State of the \gls{BT} $\btgraph{}$.
\end{definition}
\begin{definition}[\gls{BT} Relations]
  In a \gls{BT} \btdef{}, a Node $p \in \btnodes{}$ is called the \emph{parent} of a node $c \in \btnodes{}$, and $c$ a \emph{child} of $p$, if $(p, c) \in \btedges{}$.
  The set of all nodes that are children of $p$ is obtained by $\varname{children}(p, \btnodes{}, \btedges{})$.
\end{definition}
\begin{definition}[Data Graph]
The \emph{Data Graph} \datadef{}, defines data connections between Nodes of a \gls{BT}.
Its vertices are called \emph{Parameters}, and the edges \emph{Wirings}.
Section \ref{sec:data_graph} defines the \emph{Data Graph} in more detail.
\end{definition}
\begin{definition}[World State]
  In order to model interaction of a given \gls{BT} \btdef{} and Data Graph \datadef{} with the robot hardware -- including sensor readings -- and the changing values of Node States and of Parameters, we define the \emph{World State} $\btworld{} \in \btdataalph{}^x, x \in \mathbb{N}$, where $\btdataalph{}$ is an arbitrary \emph{Data Alphabet}.

  The exact alphabet used is not very important, since the functions $\varname{state}(n, \btworld{})$, $\varname{value}(p, \btworld{})$ extract the current State of a Node $n \in \btnodes{} $ and Value of a Parameter $p \in \datanodes{}$ from a given \emph{World State} $\btworld{}$, respectively.
  The following usage of these functions mostly omits the World parameter for brevity, but it is important to note that it exists, and that it (and thus, the States of Nodes and Values of Parameters) is only ever changed by the $\varname{update}$ function defined below.
\end{definition}
\begin{definition}[Nearby Robots]
  Given the World State $\btworld{}$ of a \gls{BT}, the World States of any nearby robots can be extracted from the local World State $\btworld{}$ using the function $\varname{nearbyWorlds}(\btworld{})$.
  This assumes those robots also use \glspl{BT}.
\end{definition}
\begin{definition}[\gls{BT} Environment]
  \label{def:btenv}

  The 4-tuple \btenvdef{} is called the \emph{\gls{BT} Environment}.

  The Environment can be interacted with using the $\varname{update}$ function, which describes the behavior of Nodes when \emph{ticked}, \emph{unticked}, etc.\ for a given \gls{BT} configuration and World State.
  True to its name, the function produces an updated Environment:
  \[
  \btenv{}' = \varname{update}(n, a, \btenv{})
  \]
  where $n \in \btnodes{}$ and $a \in \btactions{}$ (see definition \ref{def:node}).
\end{definition}
\begin{algorithm}[h]
  \caption{The algorithm to obtain a Subtree}
  \label{alg:subtree}

  \setstretch{1.5}
  \SetAlgoLined
  \SetAlCapSkip{2em}
  $
  \subtreenodes{} \coloneqq \{n\}
  $\;
  \While{true}{
      \tcp{Add the children of all nodes that are already in $\subtreenodes{}$}
      $
      \subtreenodes{}' \coloneqq \subtreenodes{} \cup \left( \bigcup\limits_{x \in \subtreenodes{}} \varname{children} \left( x, \btnodes{}, \btedges{} \right) \right)
      $\;
      \If{$|\subtreenodes{}'| = |\subtreenodes{}|$}{
          break\;
      }
      $
      \subtreenodes{} \coloneqq \subtreenodes{}'
      $\;
  }
  $
  \subtreeedges{} \coloneqq \left\{ (n_1, n_2) \mid (n_1, n_2) \in \btedges{} \wedge n_1,  n_2 \in \subtreenodes{} \right\}
  $\;
  \tcp{Since $\subtreenodes \subseteq \btnodes$, we can reuse $\btorderfunc{}$}
  $
  \subtreeorderfunc{} = \btorderfunc{}
  $\;
  \Return $\left( \subtreenodes{}, \subtreeedges{}, \subtreeorderfunc{} \right)$\;
\end{algorithm}
\begin{definition}[Subtree]
  \label{def:subtree}

Given the \gls{BT} \btdef{}, a \emph{Subtree} \subtreedef{} rooted at the Node $n \in \btnodes{}$ is obtained as shown in algorithm \ref{alg:subtree}.

To obtain a Subtree's Environment from the original \gls{BT}'s Environment \btenvdef{}, the Data Graph \datadef{} must also be filtered to include only those Parameters and Wirings that belong to Nodes in $\subtreenodes{}$.
  The order function $\btorderfunc{}$, Data Alphabet $\btdataalph{}$ and World State $\btworld{}$ carry over.
\end{definition}
\section{Long Running Asynchronous Tasks}
\label{sec:long_running_tasks}
Working with real-world robot systems commonly involves long-running tasks, e.g planning and following a trajectory with a robotic arm.
Classic \gls{BT} models, on the other hand, restrict nodes to execute behavior only during the \emph{tick} function.
With common tick frequencies of about 10 Hz, this becomes unfeasible for robotic use cases.
Therefore, a system for starting and stopping background tasks is introduced.

Nodes can start long-running background tasks during their \emph{tick} function.
Whenever the node is not \emph{ticked} during the current cycle, the \emph{untick} signal is given to stop all running background tasks.
The State of a Node after \emph{untick} depends on \emph{how} the background task was stopped.
If the background task was simply paused and can be resumed with the next \emph{tick}, the Node's new State is \emph{paused}.
If the background task was stopped entirely and will be \emph{restarted} with the next \emph{tick}, the Node's new State is \emph{idle}.

The resulting system has some similarities to the one in~\cite{klockner2013behavior}, but avoids the extra back and forth when a Node asks to be \emph{activated}.
Compared to~\cite{klockner2013behavior}, we make the additional assumption that \emph{idle}, \emph{success} and \emph{failure} are all \emph{rest states}.
That is, the Node must stop any background tasks before entering either of these states.
\begin{definition}[Background Tasks]
  A Node $n \in \btnodes{}$ may execute a background task if and only if $\varname{state}(n, \btworld{}) = \text{running}$.
  Upon transitioning to \textbf{any} other state, the background task must be stopped.
\end{definition}
\section{Data Graph}
\label{sec:data_graph}
\begin{figure}[hb]
  \centering
  \scalebox{0.9}{
  \begin{tikzpicture}[->,>=stealth',level/.style={sibling distance = 5cm/#1, level distance = 1.5cm}]
    \node [btnode] {$?$}
      child { \sequence{sel}
        child { node [btaction] {Detect\\ Red Ball}}
        child { node [btaction] {Pick Up\\ Red Ball}}
      }
      child { \sequence{sel}
        child { node [btaction] {Detect\\ Green Ball}}
        child { node [btaction] {Pick Up\\ Green Ball}}
      }
      ;
  \end{tikzpicture}
  }
  \caption{A Behavior Tree that will try to detect either a red or a green ball, and then proceed to pick up whichever one was detected.}
  \label{fig:concept-bt-example-2}
\end{figure}
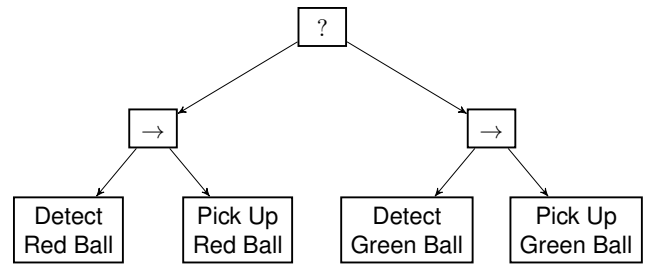

When defining the behaviors of a robot, task often depend on information acquired during the execution of their predecessors.
While some of this information can be accounted for using the control flow of a \gls{BT}, for example as seen in figure~\ref{fig:concept-bt-example-2}, this leads to two problems: redundant structures within the tree and redundant nodes with regard to functionality.
A proven solution for these issues is, allowing nodes to persist data on a shared Blackboard, which can be read by their successors.
While conceptually identical, our approach favors a data graph based solution, as it forces the explicit specification of data flows by the user, reducing the chance of runtime failures by allowing for static validity checks.
The data graph consist of \gls{BT} Nodes, parametrized with strongly typed Inputs, Outputs and Options, connected via Data Edges.
\begin{definition}[Parameters]
    Let $\btparamtypes{}$ be the set of all non-empty sets.
    Given a \gls{BT} \btdef{}, a parameter is a triple
    \begin{align*}
        p =& \left( n, \paramkind{}, \right. \left. \paramtype{} \right) \text{with:}\\
        &\text{Node}\ n \in \btnodes{}\\
        &\text{Kind}\ k \in \left\{\text{option}, \text{input}, \text{output}\right\} \\
        &\text{Type}\ t \in \btparamtypes \cup \left\{optionRef(p_{\text{ref}}) \mid p_{\text{ref}} = \left(n, \text{option}, \btparamtypes\right)\right\}
    \end{align*}
    A Parameter $p = \left( n, \paramkind{}, \paramtype{} \right)$ is called an \emph{Option}, \emph{Input} or \emph{Output} depending on its \emph{Kind} $k$. Let $\btparams$ be the set of all Parameters.
\end{definition}
The Type $\paramtype{}$ of a Parameter can be either a set from $\btparamtypes{}$ that contains the legal values for the Parameter to take (like $\mathbb{N}$ or the set of all valid Python \pythonclass{str} objects), or a \emph{Reference} to an Option, written $\varname{optionRef}(p_{\text{ref}})$.

In the latter case, $p_{\text{ref}}$ is required to have the form $(n, \text{option}, \btparamtypes{})$, where $\btparamtypes{}$ is, again, the set of all the sets that can be used as values for the Type $\paramtype{}$.
In particular, as $\btparamtypes{}$ only contains sets, it does \textbf{not} include References, so cyclic References are not possible.

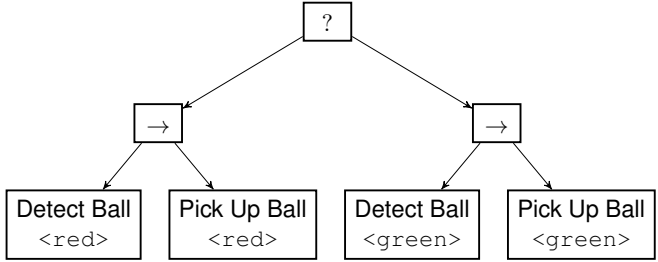
\begin{figure}[hb]
  \centering
  \scalebox{0.9}{
  \begin{tikzpicture}[->,>=stealth',level/.style={sibling distance = 5cm/#1, level distance = 1.5cm}]
    \node [btnode] {$?$}
      child { \sequence{sel}
        child { node [btaction] {Detect Ball\\\texttt{<red>\vphantom{g}}}}
        child { node [btaction] {Pick Up Ball\\ \texttt{<red>\vphantom{g}}}}
      }
      child { \sequence{sel}
        child { node [btaction] {Detect Ball\\\texttt{<green>}}}
        child { node [btaction] {Pick Up Ball\\ \texttt{<green>}}}
      }
      ;
  \end{tikzpicture}
    }
  \caption{The Behavior Tree from figure \ref{fig:concept-bt-example-2}, modified to use parameterized Nodes instead of specialized ones.}
  \label{fig:concept-bt-example-3}
\end{figure}

\begin{definition}[Parameter Values]
    \label{def:param-values}
    Given a \gls{BT} \btdef{} and World $\btworld{}$, each Parameter has a \emph{Value}:
    \begin{gather*}
        \forall p = \left( n, \paramkind{}, \paramtype{} \right) \in \btparams{} \colon
        \varname{value}(p, \btworld{}) \in \varname{resolve}(\paramtype{}, \btworld{}) \cup \{ \text{\pythonattr{None}} \}\\
        \varname{resolve} \left( \paramtype{}, \btworld{} \right) =
        \begin{cases}
            \paramtype{} &\text{ if } \paramtype{} \in \btparamtypes{} \\
            \varname{value} \left( p_{\text{ref}}, \btworld{} \right) &
            \begin{aligned}
                &\text{ if } \paramtype{} =\ \varname{optionRef} \left( p_{\text{ref}} \right) \\
                &\wedge \ p_{\text{ref}} = \left( n, \text{option}, \btparamtypes{} \right) \\
                &\wedge value \left( p_{\text{ref}}, \btworld{} \right) \neq \text{\pythonattr{None}}
            \end{aligned}
            \\
            \emptyset &\text{ else}
        \end{cases}
    \end{gather*}

    Initially, $\varname{value}(p, \btworld{}) = \text{\pythonattr{None}}$ for Inputs and Outputs.
\end{definition}
The Values of Options are static, so their initial Values are their \emph{only} Values and are defined separately.
In practice, Option Values are supplied as a dictionary to the constructor of a Node. %
\begin{definition}[Initial Option Values]
    \label{def:initial-options}
    To ensure the Types of all Parameters are known and fixed, in a \gls{BT} \btdef{} with World $\btworld{}$ there is a static set of \emph{Option Values} such that
    \[
    \forall o = \left( n, \text{option}, \paramtype{} \right) \in \btparams{} \colon \varname{value}(o, \btworld{}) = \varname{optionValue}(o, \btworld{})
    \]
\end{definition}
The \emph{Detect Green Ball} and \emph{Detect Red Ball} Nodes likely share almost all of their code, except for a single color value, and the same is true for \emph{Pick Up Red Ball} and \emph{Pick Up Green Ball}.
But Options allow us to build \emph{Detect Ball} and \emph{Pick Up Ball} nodes that can be parameterized with the color of the desired ball.
Using those, the tree could look like the one in figure \ref{fig:concept-bt-example-3}.
While this eliminates the redundant node functionality, there are still redundant structures within the tree.
\begin{definition}[Data Graph]
  Inputs and Outputs of the Nodes in a \gls{BT} \btdef{} can be connected by edges in a directed \emph{Data Graph} \datadef{}, where
  \begin{align*}
    \datanodes{} &\subseteq \btparams{}\\
    \dataedges{} &= \{ (\wiringsource{} = (n_s, \paramkind{}_s, \paramtype{}_s), \wiringtarget{} = (n_t, \paramkind{}_t, \paramkind{}_t))\ \mid \wiringsource{}, \wiringtarget{} \in \datanodes{}\\
    & \wedge \paramkind{}_s = \text{output} \\
    &\wedge \paramkind{}_t = \text{input} \\
    & \wedge \varname{resolve}(n_s, \paramtype{}_s, \btworld{}) = \varname{resolve}(n_t, \paramtype{}_t, \btworld{}) \}
  \end{align*}
\end{definition}
\begin{definition}[Data Wiring, Source, Target]
  \label{def:wirings}
  An edge $(\wiringsource{}, \wiringtarget{}) \in \dataedges{}$ is also called a \emph{Data Wiring}.
  The first element $\wiringsource{}$ is called its \emph{Source} and the second, $\wiringtarget{}$, its \emph{Target}.

  Whenever the value of the \emph{Source} $\varname{value}(\wiringsource{}, \btworld{})$ is changed, the value of the \emph{Target} $\varname{value}(\wiringtarget{}, \btworld{})$ is updated to reflect the change.
  Note that multiple Data Wirings can share the same \emph{Source} or \emph{Target}.
\end{definition}

With these definitions in place, it is now possible to refine the definition of the Data Graph of a Subtree (cf.\ def. \ref{def:subtree}):

\begin{figure}[hb]
  \centering
  \begin{tikzpicture}[->,>=stealth',level/.style={sibling distance = 5cm/#1, level distance = 1.5cm}]
    \node [btnode, bt-running] {$\rightarrow$}
      child [bt-succeeded]{ node [btnode, bt-succeeded] {$?$}
        child [bt-succeeded] { node [btaction, bt-succeeded] (detect_r) {Detect Ball\\\texttt{<red>\vphantom{g}}}}
        child [bt-idle] { node [btaction, bt-idle] (detect_g) {Detect Ball\\\texttt{<green>}}}
      }
      child [bt-running] { node [btaction, bt-running] (pickup) {Pick Up Ball}}
    ;

    \coordinate (d1) at ($ (detect_r) + (1.25,  0.5) $);

    \draw [-, color=black!75] (detect_r) edge [-, out=0, in=270] (d1);
    \draw [-, color=black!75] (d1) edge [out=90, in=180, looseness=0.5] node [auto,swap, pos=0.3, yshift=4pt, fontscale=-1] {ballPos = $(1,1)$} (pickup);
    \draw [->, color=black!75] (detect_g) edge [out=0, in=180] node [auto,swap, near start, fontscale=-1] {ballPos = \pythonattr{None}} (pickup);
  \end{tikzpicture}
  \caption[This Behavior Tree uses Data Edges to remove the need for different instances of \emph{Pick Up Ball}]{This Behavior Tree uses Data Edges to remove the need for different instances of \emph{Pick Up Ball}. It has been \emph{ticked}, and a \emph{red} ball was detected and is now being picked up.}
  \label{fig:concept-bt-example-4}
\end{figure}
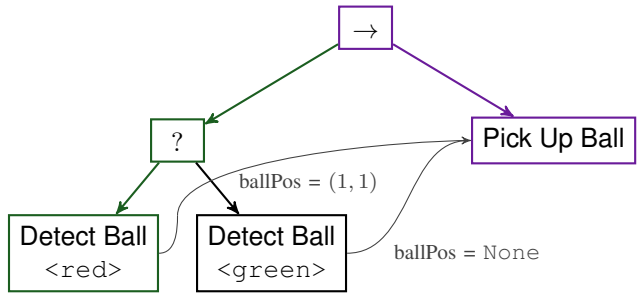

\begin{definition}[Subtree Data Graph]
  Given a \gls{BT} \btdef{}, its Data Graph \datadef{} and one of its Subtrees \subtreedef{}, the Data Graph of the Subtree $\datagraph{}'$ is defined as follows:
  \begin{align*}
    \datanodes{}' &= \left\{ \left( n, \paramkind{}, \paramtype{} \right) \mid \left( n, \paramkind{}, \paramtype{} \right) \in \datanodes{} \wedge n \in \subtreenodes{} \right\} \\
    \dataedges{}' &= \left\{ \left( \wiringsource{}, \wiringtarget{} \right) \mid \left( \wiringsource, \wiringtarget{} \right) \in \dataedges{} \wedge \wiringsource{}, \wiringtarget{} \in \datanodes{}' \right\}
  \end{align*}
\end{definition}

Unlike \gls{RAFCON}, the \gls{BT} Data Graph allows connections not just between Parameters whose Nodes have a sibling or parent-child relation, but between arbitrary Inputs and Outputs in the tree.
This avoids the problem of having to forward a piece of data multiple times through unrelated Nodes to make it available where it is needed.b

\section{Utility Functions}
\label{sec:utility_functions}
In order to automatically distribute or share capabilities between robots, a metric of the execution costs for individuals is useful to decide which robot should be chosen to execute a capability.
As we want to express capabilities of robots with \glspl{BT} we therefore introduce a utility calculation as a way to estimate the overall \emph{costs} to execute a given \gls{BT}.

\begin{definition}[Utility Function]
  In a \gls{BT} Environment $\btenv{}$, each Node is assigned a \emph{Utility Value} $\varname{utility}(n, \btworld{})$ given the World State $\btworld{}$.
  This value represents the estimated \emph{Cost} (i.e.\ lower values are better) of executing the Node using the 4-tuple $(\minsucccost{}, \maxsucccost{}, \minfailcost{}, \maxfailcost{}) \in \btutil{}$, where \btutildef{}.
  In that tuple, $\minsucccost{}$ and $\maxsucccost{}$ denote the minimum and maximum cost if the Node succeeds, and $\minfailcost{}$ and $\maxfailcost{}$ the minimum and maximum cost in case it fails.
  A cost of \notavailablemark{} in any of the four values implies that the Node cannot execute at all with the given World State $\btworld{}$, and thus implies that \textbf{all} values must be \notavailablemark{}.
  A cost of $?$ denotes the case where a Node can execute, but \emph{cannot} make an estimate of the cost of that particular case.
\end{definition}

\begin{definition}[Addition of Utility Values]
  \label{def:utility-addition}

  We define additions in \btutildef{} as follows:
  \begin{equation*}
    \forall a, b \in \mathbb{R} \cup \left[ \text{\notavailablemark{}}, ? \right] \colon
    a + b =
    \renewcommand{\arraystretch}{1.2}
    \left\{
      \begin{array}{l @{\quad} l @{\ } c @{\:} c @{\:} c @{\ } c @{\ } c @{\:} c @{\:} c}
        a + b &\text{if} &a &\in &\mathbb{R} &\wedge &b &\in &\mathbb{R}\\
        \text{\notavailablemark{}} &\text{if} &a &= &\text{\notavailablemark{}} &\vee &b &= &\text{\notavailablemark{}}\\
        ? &\text{else}\\
      \end{array}\right.
  \end{equation*}
\end{definition}
\begin{definition}[Utility Value of a \gls{BT}]
  Given a \gls{BT} Environment \btenvdef{}, the Utility Value of the Root Node $\varname{utility}(r, \btworld{})$ is called the Utility Value of the \gls{BT}.
\end{definition}

\begin{figure}[htb]
  \centering
  \begin{minipage}{0.35\textwidth}
    \centering
    \begin{tikzpicture}[->,>=stealth',level/.style={sibling distance = 2cm/#1, level distance = 1.5cm}]
      \node [btaction, bt-idle] {$\rightrightarrows$\\$<$1$>$}
      child [bt-idle] { node [btaction, bt-idle] {1}}
      child [bt-idle] { node [btaction, bt-idle] {2}}
      ;
    \end{tikzpicture}
    \caption{The example \gls{BT} used to calculate the values in table \ref{tab:concept-parallel-utility-agg}. The root succeeds if one child succeeds.}
    \label{fig:concept-parallel-util-example}
  \end{minipage}
\end{figure}
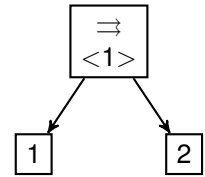

Table \ref{tab:concept-parallel-utility-agg} illustrates the utility costs for the simple \gls{BT} shown in fig. \ref{fig:concept-parallel-util-example}.
The tree succeeds if one of its childs succeeds.

\begin{table}[tb]
  \caption[Possible execution paths and Utility Values for a Parallel Node]{The possible execution paths of a Parallel Node $n_P$ with two child nodes $n_1, n_2$. In the first column, $c^i_{\left\{\text{min}, \text{max} \right\}}$ is short for the value from $\varname{utility}(n_i, \btworld{})$ corresponding to $\varname{state}(n_i, \btworld{})$. Furthermore, we assume that $\forall i \in \{1, 2 \} \colon$ $\varname{utility}(n_i, \btworld{}) =$ $(\minsucccost{} = 1, \maxsucccost{} = 10, \minfailcost{} = 2, \maxfailcost{} = 5)$.}
  \label{tab:concept-parallel-utility-agg}
  \centering
  \resizebox{\columnwidth}{!}{
  {\renewcommand{\arraystretch}{1.1}%
  \begin{tabular}{c @{\hspace{2em}} c c c c c c}
    \toprule
    Case &
           1 & 2 & 3 & 4 & 5 & 6\\
    $\varname{state}(n_1, \btworld{})$ &
        \textcolor{succeeded}{succeeded} & \textcolor{succeeded}{succeeded} & \textcolor{succeeded}{succeeded} & \textcolor{running}{running} & \textcolor{failed}{failed} & \textcolor{failed}{failed} \\
    $(c^{1}_{\text{min}}, c^{1}_{\vphantom{i}\text{max}})$ &
        $(1, 10)$ & $(1, 10)$ & $(1, 10)$ & --- & $(2, 5)$ & $(2, 5)$ \\
    $\varname{state}(n_s, \btworld{})$ &
        \textcolor{succeeded}{succeeded} & \textcolor{failed}{failed} & \textcolor{running}{running} & \textcolor{succeeded}{succeeded} & \textcolor{succeeded}{succeeded} & \textcolor{failed}{failed} \\
    $(c^{2}_{\text{min}}, c^{2}_{\vphantom{i}\text{max}})$ &
        $(1, 10)$ & $(2, 5)$ & --- & $(1, 10)$ & $(1, 10)$ & $(2, 5)$ \\ \addlinespace
    $\varname{state}(n_f, \btworld{})$ &
         \textcolor{succeeded}{succeeded} & \textcolor{succeeded}{succeeded} & \textcolor{succeeded}{succeeded} & \textcolor{succeeded}{succeeded} & \textcolor{succeeded}{succeeded} & \textcolor{failed}{failed}\\
    $(c^{P}_{\text{min}}, c^{P}_{\vphantom{i}\text{max}})$ &
         $(2, \mathbf{20})$ & $(3, 15)$ & $(\mathbf{1}, 10)$ & $(\mathbf{1}, 10)$ & $(3, 15)$ & $(\mathbf{4}, \mathbf{10})$ \\
    \bottomrule
  \end{tabular}}
}
  \[
  \varname{utility}(n_f, \btworld{}) = (\minsucccost{} = \mathbf{1}, \maxsucccost{} = \mathbf{20}, \minfailcost{} = \mathbf{4}, \maxfailcost{} = \mathbf{10})
  \]
\end{table}

\section{Shoving and Slots - Remote Execution of BTs}
\label{sec:shoving}
A core idea of the proposed approach is the transparent distribution of parts of the \glspl{BT} within the whole robot team.
The shovable decorator is introduced to enable the "shoving" of a Subtree from the current executor to other robots that are nearby and provide "Slots" to receive the expedited parts of the tree.
To decide on where to distribute a Shovable, the utility calculation is used to select the most suitable robot in the team for execution.
Results of the subtree are transparently integrated into the main \gls{BT}.

\begin{figure}[h]
  \centering
  \scalebox{0.7}{
  \begin{tikzpicture}[->,>=stealth',level/.style={sibling distance = 6cm-#1cm, level distance = 1.5cm}]
    \node [btnode] {$\rightarrow$}
      child { node [btdecorator] (shove) {Shovable}
        child { \fallback{sel}
          child { node [btaction] (detect_r) {Detect Ball\\\texttt{<red>\vphantom{g}}}}
          child { node [btaction] (detect_g) {Detect Ball\\\texttt{<green>}}}
        }
      }
      child { node [btaction] (pickup) {Pick Up Ball}}
      ;

      \coordinate (d1) at ($ (detect_r) + (2,  0.75) $);

      \draw [-, color=black!75] (detect_r) edge [-, out=0, in=180] (d1);
      \draw [-, color=black!75] (d1) edge [out=0, in=180] node [auto, near end] {ballPos} (pickup);
      \draw [->, color=black!75] (detect_g) edge [out=0, in=180] node [auto,swap] {ballPos} (pickup);

      \node[round,draw=fzi-green,label={[fzi-green] above left:Subtree Boundary}, fit=(sel)(detect_r)(detect_g)] {};

  \end{tikzpicture}
  }
  \caption[The Behavior Tree from figure \ref{fig:concept-bt-example-4}, modified to use the \emph{Shovable} Decorator]{The Behavior Tree from figure \ref{fig:concept-bt-example-4}, modified to use the \emph{Shovable} Decorator. This means the \emph{Fallback} containing the two \emph{Detect Ball} nodes can be executed remotely, if a remote executor reports a higher Utility Value than the local one. Note that the two Wirings cross the boundary between the \emph{Shovable} subtree and the rest of the \gls{BT}.}
  \label{fig:concept-bt-example-5}
\end{figure}
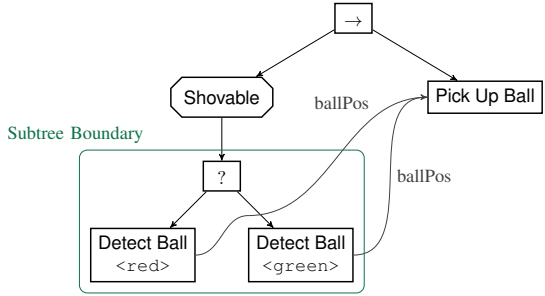

\begin{definition}[Shovable]
  \label{def:shovable}
  The \emph{Shovable} Node is a Decorator that, when ticked, extracts the Subtree Environment $\subtreeenv{}$ rooted at its only child $c$ from the Environment $\btenv{}$.
  It uses the World States of any nearby robots, $\varname{nearbyWorlds}(\btworld{})$, to determine whether to execute the Subtree locally or remotely, sending it to another robot for execution in the latter case.
\end{definition}

\begin{definition}[Public Inputs and Outputs]
  Given a \gls{BT} Environment \btenvdef{} and a Subtree \subtreedef{}, the Subtree's \emph{Public} \emph{Inputs} and \emph{Outputs} are those vertices of the Data Graph that are part of a Wiring from a Node that is part of the Subtree to one that is not:
  \begin{align*}
    \left\{ \right. \left(n, \paramkind{}, \paramtype{} \right) &\mid n \in \subtreenodes{} \wedge \left( \right.\\
    &\left( \left( \left( n, \paramkind{}, \paramtype{} \right), \left( n', \paramkind{}', \paramtype{}' \right) \right) \in \dataedges{} \wedge n' \notin \subtreenodes{} \right) \\
    &\vee \left(\left( \left(n', \paramkind{}', \paramtype{}' \right), \left(n, \paramkind{}, \paramtype{} \right) \right) \in \dataedges{} \wedge n' \notin \subtreenodes{} \right) \left.\right) \left. \right\}
  \end{align*}
\end{definition}

\begin{definition}[Slot]
  \label{def:slot}
  A \emph{Slot} $s$ is a Node that receives one remote Subtree Environment $\subtreeenv{}$ from a \emph{Shovable} Decorator at a time.
  Upon receiving a Subtree Environment, the values of its Inputs are extracted from the Subtree World State $\btworld{}'$ and encoded into the the World State $\btworld{}$ from the \emph{Slot}'s Environment \btenvdef{}.
  The $\varname{update}$ function for a \emph{Slot} then proceeds as if the Subtree was part of the \emph{Slot}'s parent \gls{BT}.

  As soon as the Subtree enters a state other than \emph{running}, the Subtree Environment is sent back to the \emph{Shovable} Decorator it was received from.
  The Environment that is sent back includes the updated World State encoding the Subtree's Node states and Parameter values.
  All information about the Subtree Environment is removed from the \emph{Slot}'s World State $\btworld{}$ when $\varname{update}(s, a, \btenv{})$ is next called after finishing execution of the Subtree.
\end{definition}

\section{IMPLEMENTATION \& DEMONSTRATION}
\label{sec:Implementation}
With the presented approach, we extended the common Behavior Tree definitions to further increase their suitabiliy for robotics applications.
Table~\ref{tab:implementations} shows how the presented approach compared to other \gls{BT} implementations available.

The addition of the graph-based parameter model provides an intuitive data sharing method between \gls{BT} nodes.
Options provide way of parametrizing nodes with static values, thus increasing node reuse.
Inputs and Outputs as a wiring based approach enforce explicit data interfaces in the nodes during build time and thus eliminate sources of errors in black board approaches  such as missing data at run time.
With all parameters being statically typed runtime errors due to mismatching types are also avoided.

To accommodate the long running tasks prevalent in robotics, we extended the Behavior Tree definitions to allow asynchronous task execution.
This ties in closely with the native integration into \gls{ROS}, where we make use of this for interacting with services and topics.

Utlitiy calculation and distributed execution with shovables present a new approach of distributing a Behavior Tree over multiple systems extending the currently available approaches towards multi robot coordination.

\subsection{ros\_bt\_py Library}
\begin{figure}[htb]
  \centering
  \includegraphics[width=1\columnwidth]{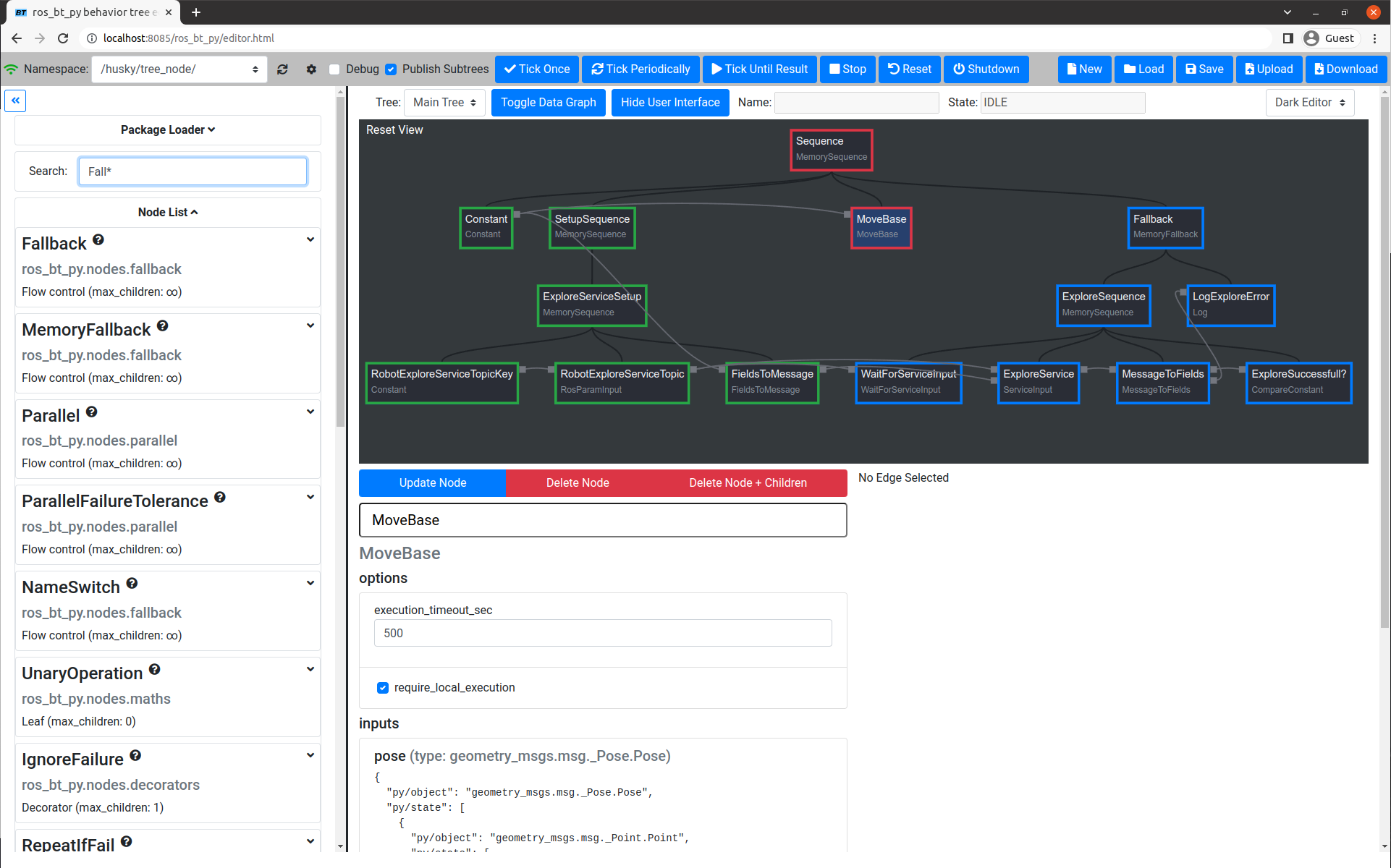}
  \caption{Screenshot of the web based GUI of the created \textbf{ros\_bt\_py} library. It can be used to create, execute and debug \glspl{BT}. Available nodes can be selected and searched for on the left, the execution options of the \gls{BT}, load, safe and debug functions and various options can be accessed via the controls at the top. The main window shows the \gls{BT} itself as well as the data connections between the nodes. The colors indicate the state of a node (green=success, yellow= active, red=failed, purple=shutdown,  blue=idle, grey=uninitialized). When a node is selected its options are shown and can be modified in the lower area. Auto completion helps the user with complex message types.}
    \label{fig:bt_gui}
\end{figure}

The concept was implemented as a python library with a native \gls{ROS} integration called \textit{ros\_bt\_py}.
In addition to the core library a web-based GUI for editing, debugging and running the \gls{BT}s is provided.
Figure\ref{fig:bt_gui} shows a screenshot of the web interface with a loaded behavior tree.
To simplify the \gls{BT} development process, the library comes with a set of common control flow and leaf nodes, as well as dedicated nodes for interacting with \gls{ROS} services, actions, topics and parameters.
Additional nodes can be implemented in python code and loaded during runtime.
The library is available as an open-source project on GitHub \footref{fn:ros_bt_py}.

\subsection{Multirobot Coordination Demonstration}

\begin{figure}[htb]
  \centering
  \includegraphics[width=0.32\columnwidth]{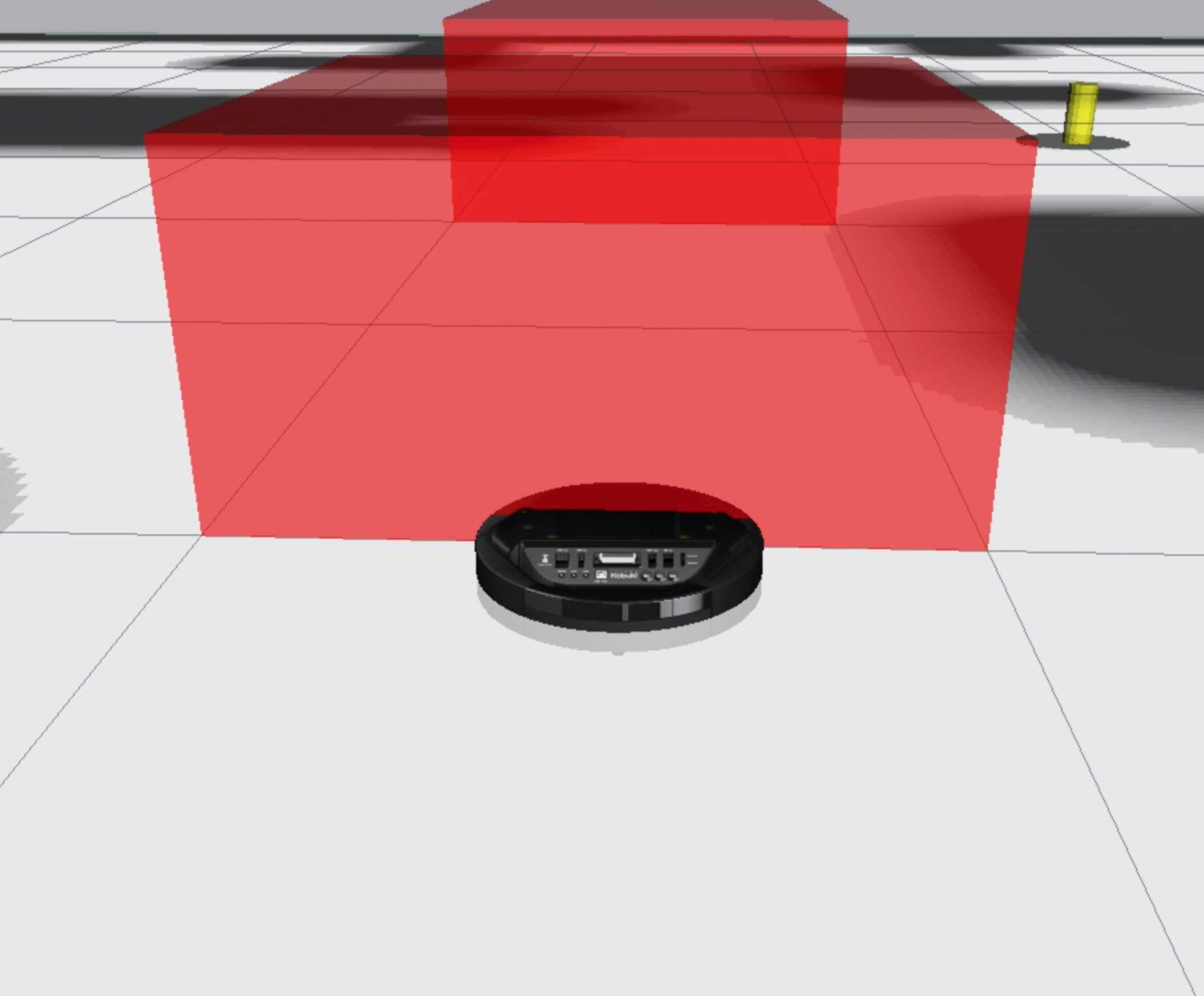}
  \includegraphics[width=0.32\columnwidth]{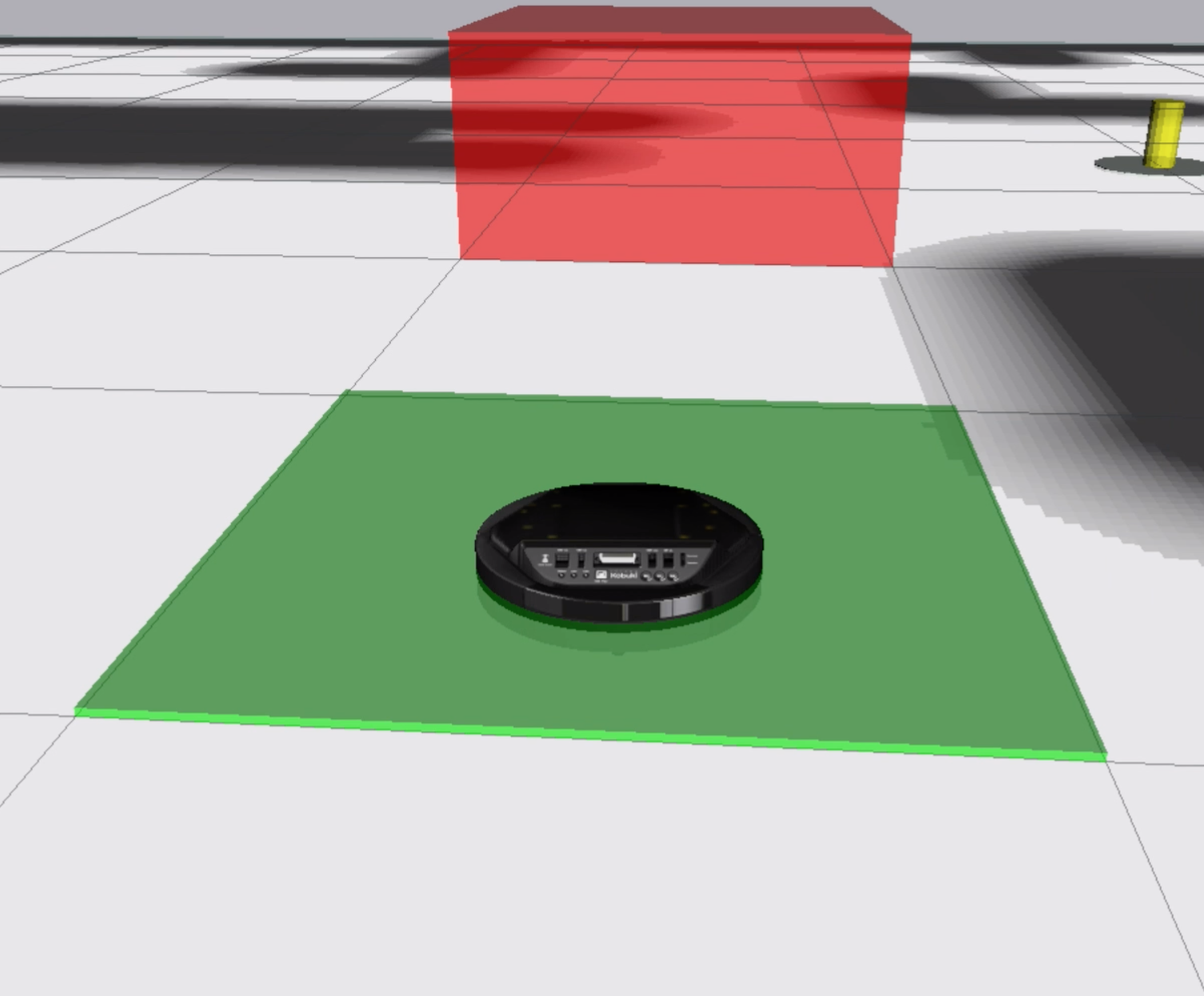}
  \includegraphics[width=0.32\columnwidth]{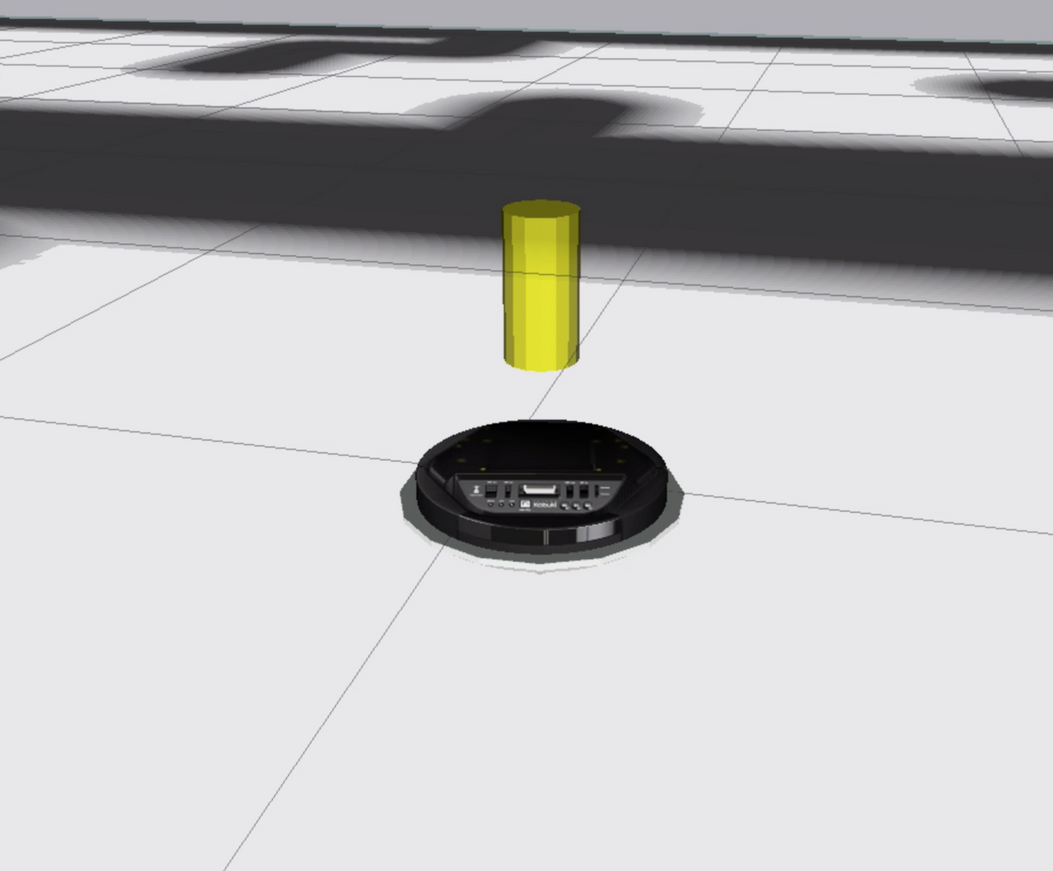}
  \caption{From left to right: Simulation visualization of a closed door, an open door and a target object.}
    \label{fig:multirobot_experiment_viz}
\end{figure}

The dynamic distribution of tasks via shovables was evaluated with the Simple Two-Dimensional Robot (STDR) simulator \cite{STDR_Simulator}.
We used a modified version of the Kobuki robot, a TurtleBot compatible robot, that was modified to support multiple instances (namespaces) and use the ground-truth localization provided by the simulator as the focus was on the mission.
Simulation helpers were implemented to provide mission functionality:
\begin{itemize}
\item{\textbf{Target Objects}}\\
Target objects can be picked up via a simple (ROS) service call.
That call takes as its arguments the ID of the object to be picked up and
the tf frame of the robot that is attempting the interaction, and returns a boolean value indicating whether the pick-up was successful.
The pick-up succeeds when the robot is close enough to the object at the time it tries to pick it up – orientation is not taken into account.
\item{\textbf{Doors}}\\
The purpose of doors is to stop robots from entering parts of the map.
In the mission, a door needs to be opened before moving past it to pick up a target object.
Using them works much like picking up a target object: The %
service request contains the door ID and robot frame, and the response informs the robot about success or failure, based on whether it was close enough to the door when it attempted to open it.
However, there is also a way to force a door open or closed, used as part of the trials to test the reactivity of the \gls{BT} system.
\end{itemize}
The Items, Doors and the states (open, closed, picked-up) are visualized in RVIZ see fig. \ref{fig:multirobot_experiment_viz}.

A single \gls{BT} was designed (fig. \ref{fig:multirobot_experiment_tree}) that first opens a simulated door, then picks up a simulated object (see  fig.\ref{fig:multirobot_experiment_viz}) via the services provied by the simulation helpers.
The Subtree responsible for opening the door is marked as shovable and is therefore automatically evaluated to be either executed on the original robot or on nearby robots that provide a RemoteTreeSlot and are able to perform the action.
The simulation \gls{BT} also showcases the use of Data Wirings and Subtree Nodes resulting in a very readable representation of the mission, and allows
for easy reuse.

Initially, all services are available to a single robot.
After starting the \gls{BT} via the editor’s Tick Until Result button, the
robot smoothly executes the mission, first opening a door by executing the OpenDoor subtree and then picking up a target object by executing the PickupObject subtree.

Now, the service for opening doors is only made available for the second robot, while the pick up objects is only available to the first robot, making it a team of heterogeneous robots.
The \gls{BT} runtime in the first robots namespace uses the same \gls{BT} as before, while second uses a simple \gls{BT} consisting of just a RemoteSlot Node providing the RemoteTreeSlot for shovables\footnote{It should be noted, that it is of course possible to provide a more complex \gls{BT}, for example with recovery or self-preservation functions.}.
After starting both \glspl{BT}, the mission is again excecuted smoothly.
The Shovable Decorator in the main \gls{BT} correctly detects the RemoteTreeSlot as a possible executor for its Subtree.
It then queries that Slot and the local environment for Utility Values.
As the services are only available to the individual robot, the binary utility values indicate that the open doors subtree can only run on the second robot.
Thus, the Subtree is shoved into the RemoteTreeSlot, which is immediately activated by the second robot’s \gls{BT}.
The second robot moves to and opens the door, after which the RemoteTreeSlot reports the successful execution of the shoved Subtree back to the main \gls{BT}.
Finally, the first robot moves to and picks up the target object.

When we force the door shut as the first robot is already executing its subtree, its execution is aborted and the tick returned to the OpenDoor subtree, which is shoved to the second robot.
Just as with a single robot case, the \gls{BT} ensures that the door remains open, commanding the second robot to re-open it if necessary due to reactive nature of the \glspl{BT}.

\subsection{Other usages}
We have successfully made use of the presented framework within several domains.
\emph{ros\_bt\_py} was used for planetary exploration \cite{schnell2018intelliRisk}, retrieval of hazardous objects with a mobile manipulator \cite{Roennau2022} coordination of multi-robot teams within the ESA Space Resources Challenge\footnote{\url{https://www.spaceresourceschallenge.esa.int/}} and in multiple reusable plug-and-play robotic componets framwork projects, namely Shop4CF\footnote{\url{https://shop4cf.eu/}} and SeRoNet\footnote{\url{https://www.seronet-projekt.de/}}.
In all cases \emph{ros\_bt\_py} could be used to quickly build complex behaviors that interact with custom systems and combine them to real world missions to great success.
The largest issues in these cases were usability issues in the GUI or missing specialized nodes which are constantly extended.

\section{CONCLUSIONS AND FUTURE WORKS}
\label{sec:conclusion}

We presented a formal Behavior Tree framework that extends previous approaches by specifically designing it for the easy use in multi robot teams.
It incorporates robotic requirements such as long running asynchronous tasks and native integration with ROS to easily interact with existing robotic systems.
By adding utility calculations and Shovables it extends its usage from a single robot to a transparent and native execution over multiple systems while considering costs.
The framework was implemented and released as open source library (\emph{ros\_bt\_py}) that increases the currently available functions of \gls{BT} implementations in a significant way.
May features, such as the type safe data edges, an easy to use web-gui or the modular and extendible node design in python enable a rapid adoption and extension of its capabilities for a wide range of use cases.
The library was evaluated in a simulation environment where the distribution and reactivity of the system could be shown and is actively used for various real world applications where it provides high level mission control of individual robots.

We are currently  extending the \gls{BT} definition with the concept of capability nodes which will allow fully dynamic task distribution and synthesis during runtime (while the tree is already ticking) as well as market based distribution methods which further support dynamic team compositions.
Furthermore, the library will be optimized for greater ease of use and will be updated to support ROS 2.

This work, forms the basis for our current approach to control and coordination of robot behavior in a flexible way for individual robots, but also to coordinate a team.
With a growing trend towards multi robot teams, we believe that such a flexible system is the right way to approach mission definition as it particularly opens up the ability to use individual capabilities of robots and therefore increases the overall team usefulness.
By providing this approach as open source library we hope that it can be used by many as a powerful tool for robot control and coordination.

\section*{ACKNOWLEDGMENT}

The research leading to these results has received funding in the IntelliRISK project under the grant agreement No. 50RA1730 by the Federal Ministry for Economic Affairs and Energy (BMWi) on the basis of a decision by the German
Bundestag and the ROBDEKON project of the German Federal Ministry of Education and Research under grant agreement No 12N14679.

\bibliographystyle{IEEEtran}
\bibliography{IEEEabrv,iros2020.bib}

\end{document}